\documentclass[10pt,twocolumn,letterpaper]{article}

\usepackage{iccv}
\usepackage{times}
\usepackage{epsfig}
\usepackage{graphicx}
\usepackage{amsmath}
\usepackage{amssymb}
\usepackage{multirow}
\usepackage[accsupp]{axessibility}

\usepackage[pagebackref=true,breaklinks=true,letterpaper=true,colorlinks,bookmarks=false]{hyperref}

\iccvfinalcopy

\begin{document}

%%%%%%%%% TITLE
\title{Decoupled Iterative Refinement Framework for Interacting Hands Reconstruction from a Single RGB Image}
\author{Pengfei Ren$^{1,2}$\quad Chao Wen$^{2}$\quad Xiaozheng Zheng$^{1,2}$\quad Zhou Xue$^{2}$\\
      Haifeng Sun$^{1}$\quad Qi Qi$^{1}$\quad Jingyu Wang$^{1}$\thanks{Corresponding authors}\quad Jianxin Liao$^{1}$\footnotemark[1]\\
$^1$State Key Laboratory of Networking and Switching Technology,  \\Beijing University of Posts and  Telecommunications \\  $^2$ PICO IDL, ByteDance, Beijing\\
{\tt\small rpf@bupt.edu.cn; \{wenchao.w, zhengxiaozheng\}@bytedance.com; xuezhou08@gmail.com;} \\
{\tt\small \{hfsun, qiqi8266, wangjingyu, liaojx\}@bupt.edu.cn}\\
}
\maketitle

%%%%%%%%% ABSTRACT
\begin{abstract}
Reconstructing interacting hands from a single RGB image is a very challenging task. On the one hand, severe mutual occlusion and similar local appearance between two hands confuse the extraction of visual features, resulting in the misalignment of estimated hand meshes and the image. On the other hand, there are complex spatial relationship between interacting hands, which significantly increases the solution space of hand poses and increases the difficulty of network learning. In this paper, we propose a decoupled iterative refinement framework to achieve pixel-alignment hand reconstruction while efficiently modeling the spatial relationship between hands. Specifically, we define two feature spaces with different characteristics, namely 2D visual feature space and 3D joint feature space. First, we obtain joint-wise features from the visual feature map and utilize a graph convolution network and a transformer to perform intra- and inter-hand information interaction in the 3D joint feature space, respectively. Then, we project the joint features with global information back into the 2D visual feature space in an obfuscation-free manner and utilize the 2D convolution for pixel-wise enhancement. By performing multiple alternate enhancements in the two feature spaces, our method can achieve an accurate and robust reconstruction of interacting hands. Our method outperforms all existing two-hand reconstruction methods by a large margin on the InterHand2.6M dataset. 
\end{abstract}

%%%%%%%%% BODY TEXT
\section{Introduction}

3D hand reconstruction plays an important role in many applications, such as virtual reality (VR), augmented reality (AR), robotics, etc. With the emergence of large-scale datasets and deep learning, single-hand pose estimation and reconstruction \cite{tompson2014real, huang2020awr, wan2018dense, moon2018v2v,  wan2019self, ren2022mining, zimmermann2017learning, cai2018weakly,mueller2018ganerated, iqbal2018hand, boukhayma20193d, zhang2019end, zimmermann2019freihand, baek2019pushing, kulon2020weakly, spurr2020weakly, baek2020weakly, moon2020deephandmesh, moon2020i2l, chen2021camera, chen2021i2uv, chen2022mobrecon} have made significant progress in the past few years. Furthermore, since two hands can express richer semantics and implement more complex operations, interacting two-hand reconstruction has received a lot of attention recently. Previous work usually relies on depth cameras \cite{ballan2012motion, oikonomidis2012tracking,kyriazis2014scalable,taylor2017articulated,mueller2019real} or multi-camera systems \cite{smith2020constraining}, which greatly limits the application scenarios of these methods. In this paper, we focus on reconstructing interacting hands from the widely available RGB image.

\begin{figure}
\begin{center}
   \includegraphics[width=0.99\linewidth]{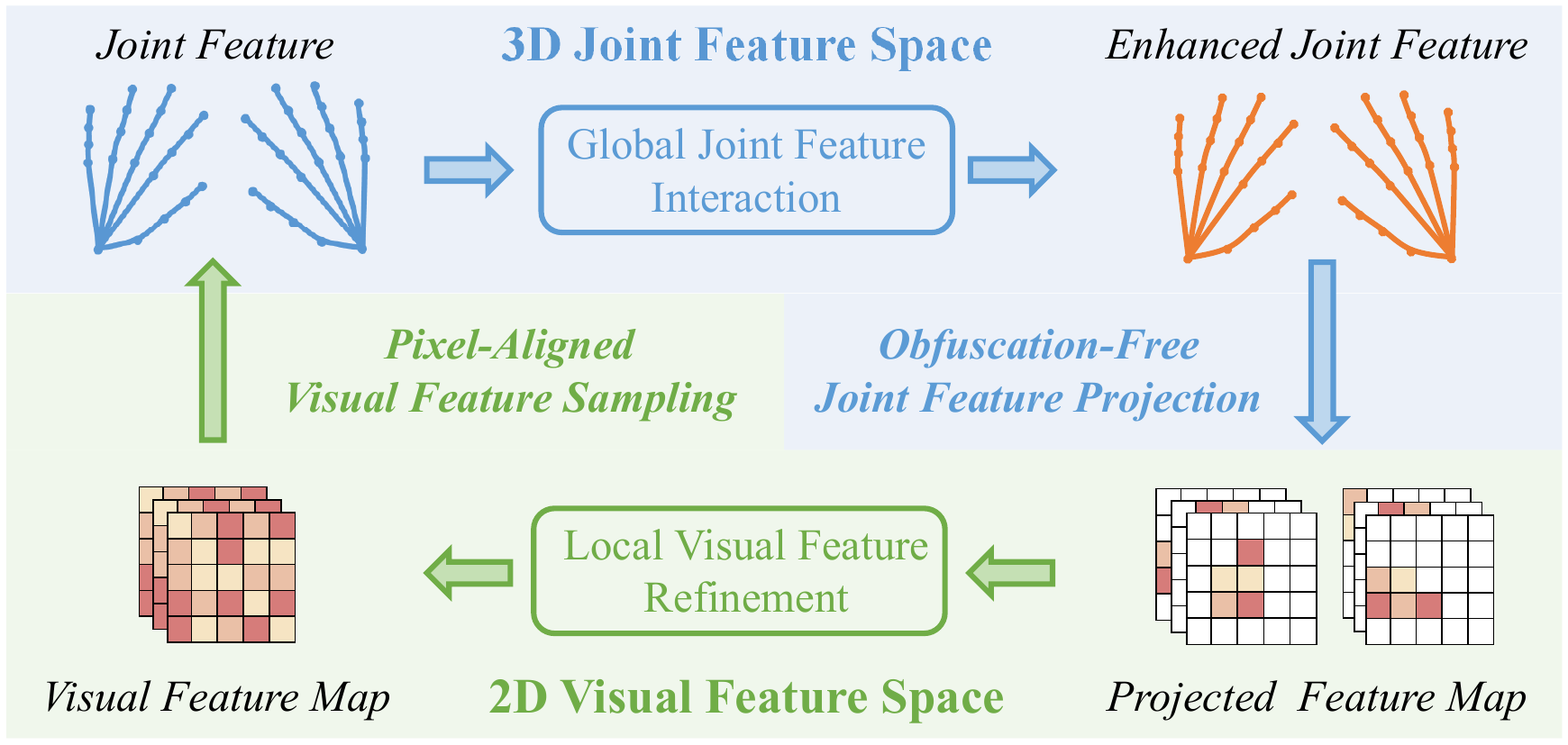}
\end{center}
   \caption{Decoupled Iterative Refinement. Our method extracts the visual feature map from the input RGB image, then iteratively performs visual feature refinement and joint feature interaction in a decoupled manner, and finally uses the enhanced joint feature for interacting hands reconstruction.}
\label{fig:algin}
\end{figure}

Compared to the single-hand reconstruction task, reconstructing the interacting hands from a single RGB image is more challenging and far from being solved. On the one hand, severe mutual occlusion between interacting hands results in a large amount of hand area being unobservable. At the same time, the self-similar appearance brings severe ambiguity and confusion to the extraction of visual representations. Thus, self-occlusion and self-similarity tend to cause misalignment between the estimated hand meshes and the input image.
On the other hand, the interacting hands have complex spatial relationships, and the dramatic increase in the degrees of freedom of the pose solution brings difficulties to the optimization of the network.

Some methods attempt to alleviate the interference of self-occlusion and similar appearance by utilizing some visual cues such as heatmaps \cite{zhang2021interacting}, joint-visibility \cite{kim2021end}, and part-segmentation \cite{fan2021learning}. Nonetheless, these methods ignore the tight dependency between interacting hands. In order to better capture the spatial relationship between hands, Hampali et al. \cite{hampali2022keypoint} extract redundant node features from the visual feature map and use a transformer to perform message passing between nodes of two hands. However, their method requires an additional association process to determine the relationship between the redundant node and the joints,  which increases the difficulty of network optimization. 
Li et al. \cite{li2022interacting} propose to perform dense interaction between the hand vertices, so as to model the spatial relationship between the hands. At the same time, they perform dense interaction between the hand vertices and pixel features, so as to achieve image-mesh alignment.
However, attention-based dense interactions are computationally expensive and suffer from the risk of overfitting.

In this paper, we decouple the difficult two-handed reconstruction task into a spatial relationship modeling problem and a pixel-level alignment problem, which can be handled in a simple but efficient way in specialized spaces, respectively.
As shown in Fig \ref{fig:algin}, our method explicitly defines two feature spaces, 2D visual feature space and 3D joint feature space. In 3D joint feature space, we represent hand information by compact joint features. We adopt a Graph Convolution Network (GCN) \cite{zhao2019semantic} and a transformer \cite{zheng20213d} to model the intra- and inter-relationships of two hands, respectively. In 2D visual feature space,  we adopt the visual feature map to represent two hands information and enhance local visual features by fusing joint features in an obfuscation-free way. We communicate the two spaces through an unambiguous 2D-3D coordinate projection relationship.
Performing long-range relational modeling in joint feature space is computationally friendly. It can take advantage of skeletal structure information, which reduces the difficulty of network optimization. At the same time, joint features with global information can provide strong disambiguation clues for local visual features, alleviating the information loss caused by self-occlusion and the ambiguity caused by self-similarity.

Experiments show that our method significantly outperforms State-Of-The-Art (SOTA) methods on the InterHand2.6M dataset \cite{li2022interacting}. At the same time, we also show qualitative images and videos (see supplementary material) of our method on multiple in-the-wild samples \cite{tzionas2016capturing,wang2020rgb2hands, bambach2015lending, shan2020understanding}, from which we can observe that our method has a strong generalization ability. Code is available at: \url{https://github.com/PengfeiRen96/DIR}.

Our contributions can be summarized as follows:
\begin{itemize}
\item We propose a decoupled iterative refinement framework for reconstructing interacting hands. Our method achieves pixel-level mesh-image alignment while efficiently modeling the spatial relationship of the hands. 

\item We model the spatial relationship of two hands through compact and semantically explicit joint nodes, which is computationally friendly and can utilize the priors of hand bone structure.

\item We propose an obfuscation-free way to project joint features into visual feature space, which alleviates the ambiguity caused by self-similarity and the absence of visual cues caused by self-occlusion.

\item Our method outperforms recent SOTA methods by a large margin and shows a strong generalization ability for the in-the-wild images.
\end{itemize}

%-------------------------------------------------------------------------
\section{Related Works}
\subsection{Single Hand Reconstruction}

Early single-hand reconstruction work relied on depth data \cite{huang2020awr, wan2018dense, moon2018v2v, tompson2014real, wan2019self, ren2022mining}, but with the emergence of large-scale datasets and the development of deep learning, RGB-based hand reconstruction has made great progress. Pioneering RGB-based work focus on only estimating hand pose from input \cite{zimmermann2017learning, spurr2018cross, iqbal2018hand, cai2018weakly}. With the proposal of parametric hand models, some work \cite{zhou2020monocular, zhang2019end, boukhayma20193d, Moon_2022_CVPR, kong2022identity} attempt to reconstruct hands directly from RGB images using MANO models. However, it is challenging to predict the parameters of the hand model from a single RGB image, which leads to difficulties in network optimization and is prone to mesh-image misalignment \cite{tang2021towards}. To alleviate these problems, some works propose to use GCN \cite{ge20193d, kulon2020weakly, choi2020pose2mesh} or transformer \cite{lin2021mesh, Lin_2021_CVPR, cho2022cross} to directly estimate the coordinates of the vertices of the hand mesh. However, reconstructing the hand without prior knowledge can easily lead to the collapse of the predicted mesh, even if these methods adopt some constraint terms to keep the generated mesh smooth. Therefore, in order to keep the estimated hand model reasonable, we adopt a parametric hand model and alleviate the mesh-image misalignment through the iterative 3D spatial relationship modeling and the 2D feature enhancement.

\subsection{Interacting Hand Reconstruction}
Interacting hand reconstruction is a very challenging problem. Some pioneering work \cite{ballan2012motion, oikonomidis2012tracking,kyriazis2014scalable} fit a parametric hand model with observed depth data by optimizing an energy function. 
These methods tend to be trapped in local optima and are computationally expensive. 
A common approach is to train a deep neural network to predict some visual cues, such as segmentation\cite{taylor2017articulated,mueller2019real}, pixel-mesh correspondence map \cite{mueller2019real,wang2020rgb2hands} or the dense relative depth map of the interacting hands \cite{wang2020rgb2hands}, to reduce the search space of hand poses and the optimization difficulty of the energy function. However, this hybrid approach cannot be trained in an end-to-end manner and the optimization process may still fall into local minima. 
Recently, Smith et al. \cite{smith2020constraining} propose a multi-view camera system, which can reconstruct high-fidelity interacting hand motions. However, this method requires custom-built dedicated hardware and the algorithm is time-consuming.

In recent years, with the proposal of the large-scale interacting hand dataset InterHand2.6M \cite{moon2020interhand2}, great progress has been made in single RGB-based 3D interacting hands reconstruction. Moon et al. \cite{moon2020interhand2} extend the single-hand pose estimation method to the two-hand interacting scenario, which predicts the 2.5D heatmap of the two hands simultaneously. 
Some works improve the accuracy of the interacting hand estimation by incorporating some visual cues such as joint-visibility \cite{kim2021end} and part-segmentation \cite{fan2021learning}. Rong et al. \cite{rong2021monocular} propose a two-stage framework to alleviate the collision problem between the hands. In order to reduce the mutual interference between interacting hands, Zhang et al. \cite{zhang2021interacting} propose to use heatmaps to make the network focus on specific regions, and Meng et al. \cite{meng2022hdr} adopt an erase mechanism to convert the two-hand image into two single-hand images. However, these methods do not adequately model the dependencies between the two hands. Hampali et al. \cite{hampali2022keypoint} adopt the transformer to model the interaction between two hands, which is robust but still hard to mitigate misalignment between the estimated hand pose and the image. Li et al. \cite{li2022interacting} progressively enhance mesh vertex features with image features while performing information interaction between two-hand meshes, which is helpful for mesh-image alignment. However, performing dense mesh-mesh and mesh-images interactions is computationally complex and prone to overfitting.

\subsection{Pixel-level Alignment}
Estimating pixel-aligned 3D mesh directly from a single RGB image is quite challenging, either by estimating parametric models or by directly predicting mesh vertex coordinates. Wang et al. \cite{wang2018pixel2mesh} and Wen et al. \cite{wen2019pixel2mesh++} propose to use the camera intrinsic matrix to obtain the perceptual features of each 3D mesh vertex from the 2D image features according to the 3D-2D coordinate relationship. Further, Zhang et al. \cite{zhang2021pymaf} and Tang et al. \cite{tang2021towards} extend this strategy to extract human body mesh and hand mesh features from visual features, respectively, in order to obtain a more accurate mesh-image alignment. In addition to sampling mesh features using coordinate relations, another way \cite{lin2021mesh,li2022interacting} is to use a transformer to perform densely interaction between image features and vertex features.  However, this method is computationally expensive and relatively sensitive to self-occlusion (the correctness of the interaction relationship is affected by the quality of the feature itself). In particular, the above two methods only perform feature enhancement in the 3D vertex space, ignoring the role of the 2D feature space. Specifically, the local receptive field mechanism of 2D convolution operation provides intrinsic inductive bias, which can efficiently and effectively utilize local features for pixel-level refinement. Our method projects joint features with global information back into the 2D visual feature space, which further provides strong cues for 2D convolution to achieve more accurate pixel-level alignment.

\begin{figure*}
\begin{center}
   \includegraphics[width=0.99\linewidth]{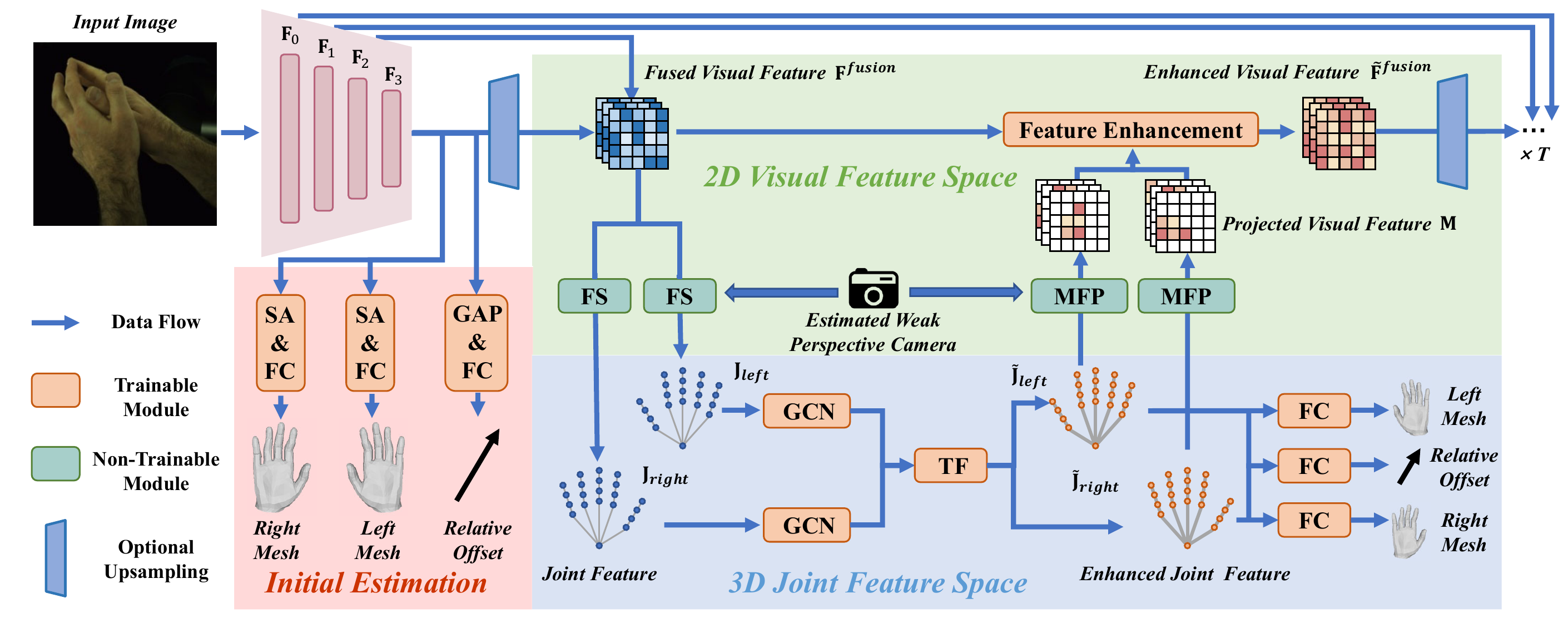}
\end{center}
   \caption{Our framework. 
   We utilize the global features extracted by the encoder to predict the initial hand meshes and the relative offset of the hands. 
   Then, the decoder gradually fuses the multi-scale visual feature maps from the encoder and refines the hand meshes and the relative offset. 
   `SA' and `GAP' represent spatial attention and global average pooling, respectively.
   `FS' and `MFP' represent joint-wise feature sampling and multi-plane feature projecting, respectively.
   `TF' represents a multi-layer transformer. In particular, since our method limits the resolution of feature maps to a maximum of 32$\times$32 during decoding, we only need to adopt the upsampling in the first refinement stage. In addition, for feature maps from the encoder with a resolution greater than 32$\times$32, we change its resolution by downsampling. 
   }
\label{fig:network}
\end{figure*}

\section{Method}
In this paper, we propose a decoupled iterative refinement module for interacting two-hand reconstruction from a single RGB image. As shown in Fig. \ref{fig:network}, we adopt an encoding-decoding network structure. The encoder extracts multi-scale visual features from the input image, and uses global features to estimate the initial hand meshes and the relative offset of two hands. Then, the decoder progressively enhance the visual feature maps and refine the hand meshes and the relative offset. During the decoding process, we iteratively perform two-hands spatial relationship modeling and visual feature refinement in a decoupled manner.

\subsection{Encoder and Initial Estimation}
We adopt a ResNet-50 \cite{He_2016_CVPR} pretrained on ImageNet \cite{deng2009imagenet} as the encoder, from which we can obtain multi-scale visual features $\{\mathbf{F}_{n} \in \mathbb{R}^{C_{n} \times H_{n}\times W_{n}}\}_{n=0}^{n=N-1}$, where $N,C_{n},H_{n},W_{n}$ represent the number of visual feature scales, the channel dimension, height and width of the $n$-th feature map respectively. In general, direct regression of vertex coordinates can achieve higher accuracy mesh prediction \cite{tang2021towards, lin2021mesh, ge20193d, kulon2020weakly}. However, this method is prone to generate collapsed and unreasonable hand mesh, so the robustness is relatively poor. Therefore, we use the global features extracted by the encoder to regress the parameters of a parameterized hand model MANO \cite{MANO} directly and then improve the accuracy of the initial meshes through subsequent iterative refinement. 
In particular, the two hands should have their own unique features, so we adopt a simple and lightweight attention module to separate the features of the two hands from the highest-level image feature map $\mathbf{F}_{N-1}$. Taking the left hand as an example, the left hand global feature $\mathbf{G}_{left} \in \mathbb{R}^{C_{N-1}}$ can be obtained by an attention map $\mathbf{A}_{left} \in \mathbb{R}^{1 \times H_{N-1}\times W_{N-1}} $ as follow:
\begin{equation}
   \mathbf{A}_{left} =  Sigmoid\left ( Conv_{left} \left ( \mathbf{F}_{N-1} \right )  \right ),
  \label{eq:hand_attn}
\end{equation}
\begin{equation}
   \mathbf{G}_{left} = AvgPool\left ( \mathbf{A}_{left} * \mathbf{F}_{N-1}  \right ). 
  \label{eq:hand_feat}
\end{equation}
Then, we estimate model parameters ($\mathbf{\theta}_{left} \in \mathbb{R}^{62}$ and $\mathbf{\theta}_{right}\in \mathbb{R}^{62}$) and weak perspective camera parameters ($\mathbf{P}_{left}\in \mathbb{R}^{3}$ and $\mathbf{P}_{right}\in \mathbb{R}^{3}$) of the two hands from $\mathbf{G}_{left}$ and $\mathbf{G}_{right}$ through Fully Connected (FC) layers, respectively. At the same time, we predict the relative offset $\mathbf{O} \in \mathbb{R}^{3}$ of the two hands from the features $\mathbf{F}_{N-1}$ through a 2D average pooling and a FC layer.

\subsection{Decoder and Iterative Refinement}
Modeling the 3D spatial relationship of the two hands and aligning the estimated mesh with the observed 2D image are two major challenges for interacting hands reconstruction. We address these two problems in the 2D image feature space and 3D joint feature space, respectively, in a decoupled manner. 
In a single refinement stage, first, we extract joint features from the 2D image feature map according to the 3D-to-2D coordinate relationship. Then, in the 3D joint feature space, we perform intra- and inter-hand information interaction to capture the complex spatial dependencies between two hand joints. Finally, we project the joint-wise features with global context information back to the 2D image space in an unobfuscated way, which provides strong disambiguation clues for local visual features refinement. In particular, we take a total of $T$ refinement stages.

\subsubsection{Constructing Joint Feature}
For the $t$-th refinement stage, given the feature map from the previous decoding layer and the skipped image feature map $\mathbf{F}_{N-t-2}$ from the corresponding encoder layer, we concatenate them together and obtain the fused feature map $\mathbf{F}^{fusion}$ by a 1$\times$1 convolution layer. 
Then, similar to \cite{tang2021towards, zhang2021pymaf, wang2018pixel2mesh, qiu2020peeking, Moon_2022_CVPR}, we obtain joint-wise visual feature $\mathbf{J}^{visual} \in \mathbb{R}^{C \times J}$ from $\mathbf{F}^{fusion}$ via a 3D-to-2D coordinate projection and a bilinear interpolation around each projected joint coordinate, where $J, C$ represent the joint number and channel dimension of the joint feature.
The 3D hand joint coordinates are obtained from the 3D hand mesh predicted in the previous stage.
In particular, the projection here is determined by the estimated weak perspective camera, so our method does not require camera intrinsics.
In addition, we encode the estimated joint coordinates into coordinate features $\mathbf{J}^{coord} \in \mathbb{R}^{C \times J}$ through a FC layer. With the joint-wise visual feature and coordinate features, we can obtain the initial joint features $\mathbf{J} = \mathbf{J}^{coord} + \mathbf{J}^{visual}$. We perform joint feature extraction for two hands independently. In particular, we use the predicted relative offset $\mathbf{O}$ to move the left-hand and right-hand coordinates in the same 3D space. 

\subsubsection{Modeling Spatial Relationship in 3D Space}
For the 3D spatial relationships, our method mainly focuses on two parts, one is the joint dependencies of a single hand, and the other is the spatial context relationships between two hands. 
First, there are explicit dependencies between the joints of a single hand. Utilizing the intrinsic dependency of the hand structure to perform information interaction between joints can reduce the difficulty of network optimization and alleviate the interference of low-quality features.
For example, when a joint lacks explicit visual cues due to occlusion, we can infer its location based on its related joints such as its parent and child joints. Therefore, we utilize a GCN \cite{zhao2019semantic} to perform intra-information interaction between the joint nodes of a single hand based on the skeletal structure.
Meanwhile, for the tightly interacting hands, there are more complex and flexible spatial relationships between the joints of the two hands. Therefore, we adopt a multi-layer transformer \cite{zheng20213d} to model the relationship between two-hand joints. By using the GCN and the transformer for information interaction, we can obtain enhanced joint features $\mathbf{\widetilde{J}} \in \mathbb{R}^{C \times J}$ with global information, which are used for MANO parameter prediction and subsequent visual feature enhancement.

Joint nodes have clear semantics, which can take advantage of hand bone structure during interaction and reduces the optimization difficulty of the network. Compared with using redundant nodes to model the two-hand relationship \cite{hampali2022keypoint}, our method can avoid the extra node-to-joint assignment. Compared with performing information interaction between the mesh vertices of two hands \cite{li2022interacting}, our method is computationally efficient and can avoid overfitting.

\subsubsection{Enhancing Visual Feature in 2D Space}

Benefiting from the local receptive field mechanism and intrinsic inductive bias, 2D convolution can capture local structures efficiently and effectively. 
However, convolution operations are difficult to model long-range relationships, even when stacking multiple convolutional layers \cite{luo2016understanding}. Therefore, 2D visual features are susceptible to self-occlusion or self-similar appearance interference. 
To alleviate this problem, we project joint features with global information back to 2D image features, which can provide strong disambiguation clues to local visual features. Furthermore, in order to prevent feature confusion when different joint features are projected back to the same or near pixel positions, we propose a Multi-plane Feature Projecting (MFP) mechanism.
Specifically, we independently project each joint feature to a feature map $\mathbf{M}_{j} \in \mathbb{R}^{C \times H \times W}$ and then concatenate $2J$ feature maps as the final projected feature map $\mathbf{M} \in \mathbb{R}^{(2J \times C)\times H \times W}$. 
Previous work \cite{li2022interacting, tang2021towards, zhang2021pymaf, wang2018pixel2mesh} focusing on mesh-image alignment only extracts vertex features unidirectionally from 2D image features and performs refinement in 3D mesh space. These methods do not enhance visual features and ignore the role of pixel-level refinement. 

\subsection{Loss Functions} 
The loss function consists of three parts, including a MANO loss, an offset loss, and a pixel-wise loss.

\textbf{MANO Loss.}  We supervise the hand joints and meshes predicted by the network. Similar to previous methods \cite{li2022interacting}, we supervise the root-relative 3D joint coordinates $\mathbf{C}^{3D}$, 2D joint coordinates $\mathbf{C}^{2D}$, 3D coordinates of mesh vertices $\mathbf{V}^{3D}$, and 2D coordinates of mesh vertices $\mathbf{V}^{2D}$ as follows:
\begin{equation}
   L_{joint} = \sum_{i=0}^{T-1} \sum_{j=0}^{J-1} L1\left ( \mathbf{C}^{3D}_{i,j} ,\mathbf{C}^{3D, gt}_{i,j} \right ) +  L1\left ( \mathbf{C}^{2D}_{i,j} ,\mathbf{C}^{2D, gt}_{i,j} \right ) ,
  \label{eq:loss_joint}
\end{equation}
\begin{equation}
   L_{mesh} = \sum_{i=0}^{T-1} \sum_{j=0}^{V-1} L1\left ( \mathbf{V}^{3D}_{i,j} ,\mathbf{V}^{3D, gt}_{i,j} \right ) +  L1\left ( \mathbf{V}^{2D}_{i,j} ,\mathbf{V}^{2D, gt}_{i,j} \right ) ,
  \label{eq:loss_mesh}
\end{equation}
where $L1$ represents the smooth L1 loss \cite{girshick2015fast, ren2019srn}; $T,J,V$ represents the number of iterative refinements, the number of joints and the number of mesh vertices, respectively. Besides, similar to previous methods \cite{li2022interacting, tang2021towards, moon2020i2l}, we also adopt a normal consistency loss and an edge length consistency to maintain the smoothness of the estimated mesh.

\textbf{Offset Loss.} Accurately estimating the offset between the two hands is important for modeling the spatial relationship of the hands. Therefore, we supervise the offsets of the joints of two hands.
\begin{equation}
L_{offset}= \sum_{i=0}^{T-1} L1(\mathbf{O}_{i}, \mathbf{O}^{gt}_{i}) 
\end{equation}

\textbf{Pixel-wise Loss.} Similar to \cite{zhang2021pymaf}, we utilize auxiliary tasks for pixel-level supervision, enhancing the reliability of visual features. Specifically, we predict two-hand segmentation and dense correspondence maps \cite{li2022interacting, mueller2019real, wang2020rgb2hands}, and supervise them with mean square error (MSE) loss.

\section{Experiment}

\subsection{Implementation Details}
We train and evaluate our method on a single server with an NVIDIA A100 Tensor Core GPU. The network is implemented within PyTorch. We train our network using the AdamW \cite{loshchilov2017decoupled} optimizer with an initial learning rate of 3e-4 and a cosine decay learning rate schedule \cite{loshchilov2016sgdr}. The whole training process takes 50 epochs with a batch size of 64. We perform data augmentation including random rotation, random scaling, random translation, random horizontal flipping and motion blur. We crop out the hand region based on the 2D coordinates of the hand vertices and resize it to 256 $\times$ 256. 
More details about network structure and training details are provided in the supplementary material.

\subsection{Datasets}

\textbf{InterHand2.6M.} We majorly conduct experiments on InterHand2.6M \cite{moon2020interhand2}, which provides multi-view RGB images with two-hand mesh and joint 3D annotation. Instead of using multi-view information, we treat all images as single-view images. This dataset is very challenging, it contains complex two-hand interaction poses and covers large-scale perspective changes. 
Following the practice of \cite{li2022interacting}, we use the 5 FPS version of the released data and only use the interacting two-hand data. Specifically, it consists of 366K training images and 261K testing images.  

\textbf{In-the-wild Datasets.} We conduct qualitative experiments on RGB2Hands dataset \cite{wang2020rgb2hands}, EgoHands dataset \cite{bambach2015lending}, 100DOH dataset \cite{shan2020understanding} and the dataset proposed by Tzionas et al. \cite{tzionas2016capturing}. These datasets have complex interacting hand samples, diverse backgrounds, realistic lighting conditions and varying image quality, which can provide a comprehensive evaluation of the generalization ability of our approach.

\subsection{Evaluation Metrics}
First, we adopt Mean Per Joint Position Error (MPJPE) and Mean Per Vertex Position Error (MPVPE) to measure the accuracy of the pose and shape of the estimated hand meshes. For a fair comparison, we follow the previous work \cite{moon2020interhand2, hampali2022keypoint} that use the wrist as root to perform joint alignment and scale the prediction according to the ground-truth bone length when evaluating. In particular, in qualitative experiments, we do not perform root joint alignment and scaling. second, we adopt the Mean Relative-Root Position Error (MRRPE) to evaluate the root-relative the translation between the hands. Third, to evaluate the Mesh-image Alignment Accuracy (MIAA), we calculates the 2D distance in image pixels between the projected ground truth vertices and the predicted vertices. In addition, we report the Percentage of Correct Keypoints (PCK) and Area Under the Curve (AUC) between 0 and 50 millimeters.

\begin{table}[t]
\begin{center}
\small
\setlength\tabcolsep{2pt} 
\begin{tabular}{l|c|cc|ccc|ccc}
\hline
\multicolumn{1}{c|}{ID} & IR                       & GCN                 & TF                  & SFP     & MHP     & MFP     & MPJPE & MPVPE & MIAA \\ \hline
1                       &                          &                     &                     &         &         &         & 12.44 & 12.11 & 7.41                                                 \\ \hline
2                       & $\surd$                  &                     &                     &         &         &         & 11.30 & 11.03 & 6.60                                                 \\ \hline
3                       & \multirow{3}{*}{$\surd$} & $\surd$             &                     &         &         &         & 10.94 & 10.75 & 6.42                                                 \\
4                       &                          &                     & $\surd$             &         &         &         & 10.98 & 10.80 & 6.45                                                 \\
5                       &                          & $\surd$             & $\surd$             &         &         &         & 10.73 & 10.53 & 6.31                                                 \\ \hline
6                       & \multirow{3}{*}{$\surd$} & \multicolumn{2}{c|}{\multirow{3}{*}{ALL}} & $\surd$ &         &         & 10.68 & 10.45 & 6.27                                                 \\
7                       &                          & \multicolumn{2}{c|}{}                     &         & $\surd$ &         & 10.65 & 10.44 & 6.27                                                 \\
8                       &                          & \multicolumn{2}{c|}{}                     &         &         & $\surd$ & 10.49 & 10.26 & 6.18                                                 \\ \hline
\end{tabular}
\end{center}
\caption{We report the MPJPE (mm), MPVPE (mm) and MIAA (pixel) on InterHand2.6M dataset. `AM' represents using attention map to split features of the left and right hands. `IR' stands for adopt iterative refinement. `SFP' and `MHP' represent the single-plane feature projection and the multi-plane heatmap projection.}
\label{table:ablation}
\end{table}

\subsection{Ablation Study}

\textbf{Basic Network.} In this section, we experiment with different ways of predicting initial MANO parameters. Here, our network does not adopt iterative refinements or decoders. As shown in Table \ref{table:ablation}, the basic method (ID 1) regresses the MANO parameters and the relative offset directly, which performs poorly. Similar to \cite{zhang2021pymaf}, we extract pixel-aligned features from visual features based on joint coordinates and perform two iterations of correction (ID 2), which can significantly improve the performance of the network. In subsequent experiments, we adopt two refinement stages, by default.

\begin{figure}
\begin{center}
   \includegraphics[width=0.99\linewidth]{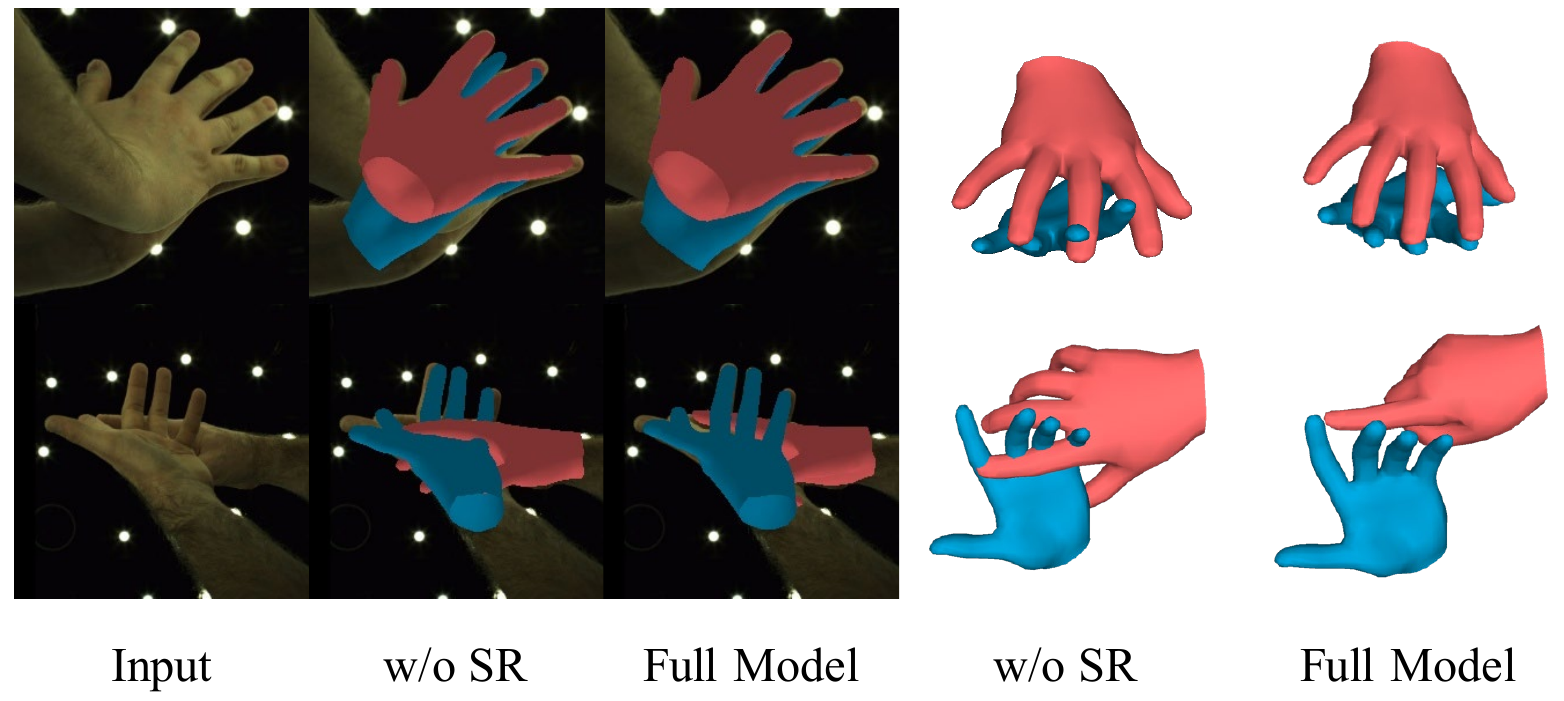}
\end{center}
   \caption{Qualitative ablation study. `w/o SR' means removing the GCN and the transformer from our full model. In addition to the mesh that overlaps the input image, we also show the estimated two-hand mesh from the other viewpoint. }
\label{fig:inter}
\end{figure}

\textbf{Modeling Spatial Relationships} In this section, we evaluate the impact of modeling intra- and inter-hand relationships on performance.  Compared with not modeling any spatial relationship (ID 2), either using GCN (ID 3) for information interaction between single-hand joints or using a transformer (ID 4) for spatial relationships modeling between two-hand joints can bring a significant performance improvement. Furthermore, using GCN and transformer at the same time (ID 5) can achieve the best performance. 

In Fig. \ref{fig:inter}, we show the importance of Spatial Relationship (SR) modeling between hands. On the one hand, abandoning spatial interactions of the two hands leads to severe intersections and collisions, especially for invisible parts (row1). On the other hand, for the samples with visible interaction regions but complex interaction pose, abandoning two-hand information interaction also leads to the wrong spatial relationship between interacting joints (row2).

\begin{figure}[t]
\begin{center}
   \includegraphics[width=0.99\linewidth]{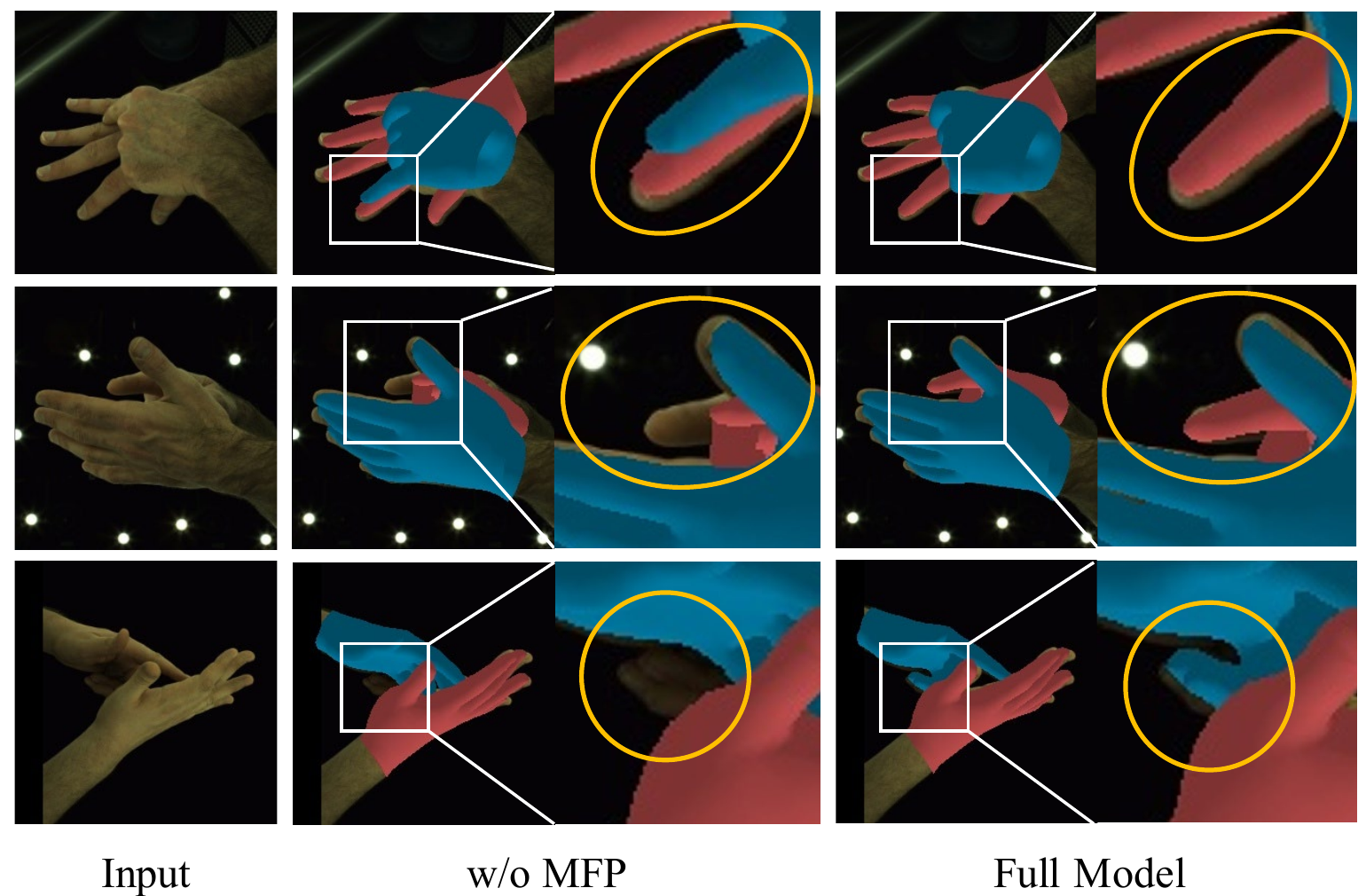}
\end{center}
  \caption{Qualitative Ablation study. `w/o MFP' means removing the MFP process from our full model. }
\label{fig:proj}
\end{figure}

\textbf{Joint Feature Projection} Joint features with global information can provide strong disambiguation cues for local visual features. First, if all joint features are projected into a single plane (ID 6), the features of different joints will be confused, so the performance also drops to a certain extent compared with ID 8. Converting the joint coordinates into multiple heatmaps (ID 7) avoids the confusion of different joints, but the information passed to the visual features is limited (only position information), so the performance is not good enough. Adopting MFP (ID 8) not only avoids feature confusion but also preserves the information of node features well, which achieves the best performance.

In Fig. \ref{fig:proj}, we show the effect of MFP. First, we observe that adopting MFP can alleviate the ambiguity caused by the self-similarity between hands (row1), which indicates that the projected two-hand information provides strong disambiguation cues for local visual features. Second, when there is severe slef-occlusion between hands, the projected information can enhance the local visual feature of unobservable regions, significantly improving the robustness of the estimation for the occluded hand (row2). We also observe that adopting MFP can help the network to focus on some areas that are easily ignored, such as dark regions (row3).

\begin{figure}[t]
\begin{center}
   \includegraphics[width=0.99\linewidth]{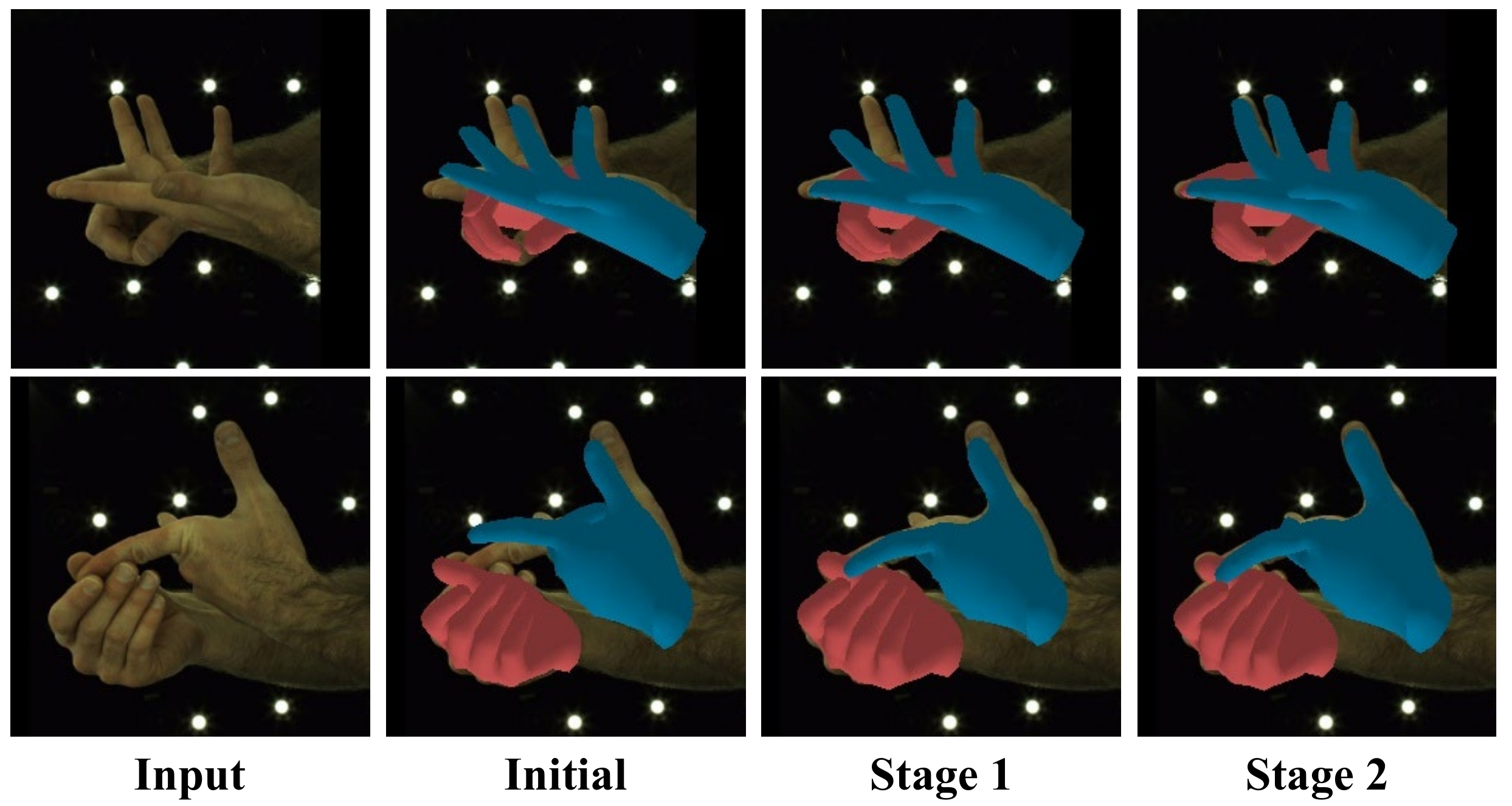}
\end{center}
  \caption{From left to right, we show that the initial hand meshes are gradually corrected. A more accurate spatial relationship modeling can provide stronger disambiguation cues for 2D feature refinement in 2D visual space; better 2D visual features help to construct more representative joint nodes in 3D joint space.}
\label{fig:refine}
\end{figure}

\textbf{Iterative Refinement} In this section, we explore the effect of the number of refinement stages on the performance. With one, two, and three refinement stages, MPJPE can reach 10.95 mm, 10.49 mm, and 10.44 mm, respectively, which are about 12.0\%, 15.7\%, and 16.1\% higher than the initial estimation. A better balance of speed and accuracy can be achieved with two-stage refinement (31 FPS), while higher performance can be achieved with three-stage refinement (23 FPS). As shown in Fig \ref{fig:refine}, our method can effectively correct initial estimates with misaligned meshes or wrong spatial relationships between hands.

\begin{table}[tbp]
\begin{center}
\setlength\tabcolsep{5pt} 
\begin{tabular}{l|cccc}
\hline
Method            & MPJPE & MPVPE & MIAA & MRRPE \\ \hline
InterNet  \cite{moon2020interhand2}        & 14.42  &  - & - &  30.08 \\
InterShape \cite{zhang2021interacting}     & 12.54  & 12.26 & 7.40 & -\\
KPT \cite{hampali2022keypoint}     & 12.42  & -  & - & 29.17 \\
IntagHand  \cite{li2022interacting}        & 12.40  & 12.09 & 7.11 & 30.40 \\ 
\hline
Ours              & \textbf{10.49}   & \textbf{10.26} & \textbf{6.18} & \textbf{28.98} \\ \hline
\end{tabular}
\end{center}
\caption{We report MPJPE (mm), MPVPE (mm), MIAA (pixel) and MRRPE (mm) on InterHand2.6M.}
\label{table:sota}
\end{table}

\begin{figure}[tbp]
\begin{center}
   \includegraphics[width=0.99\linewidth]{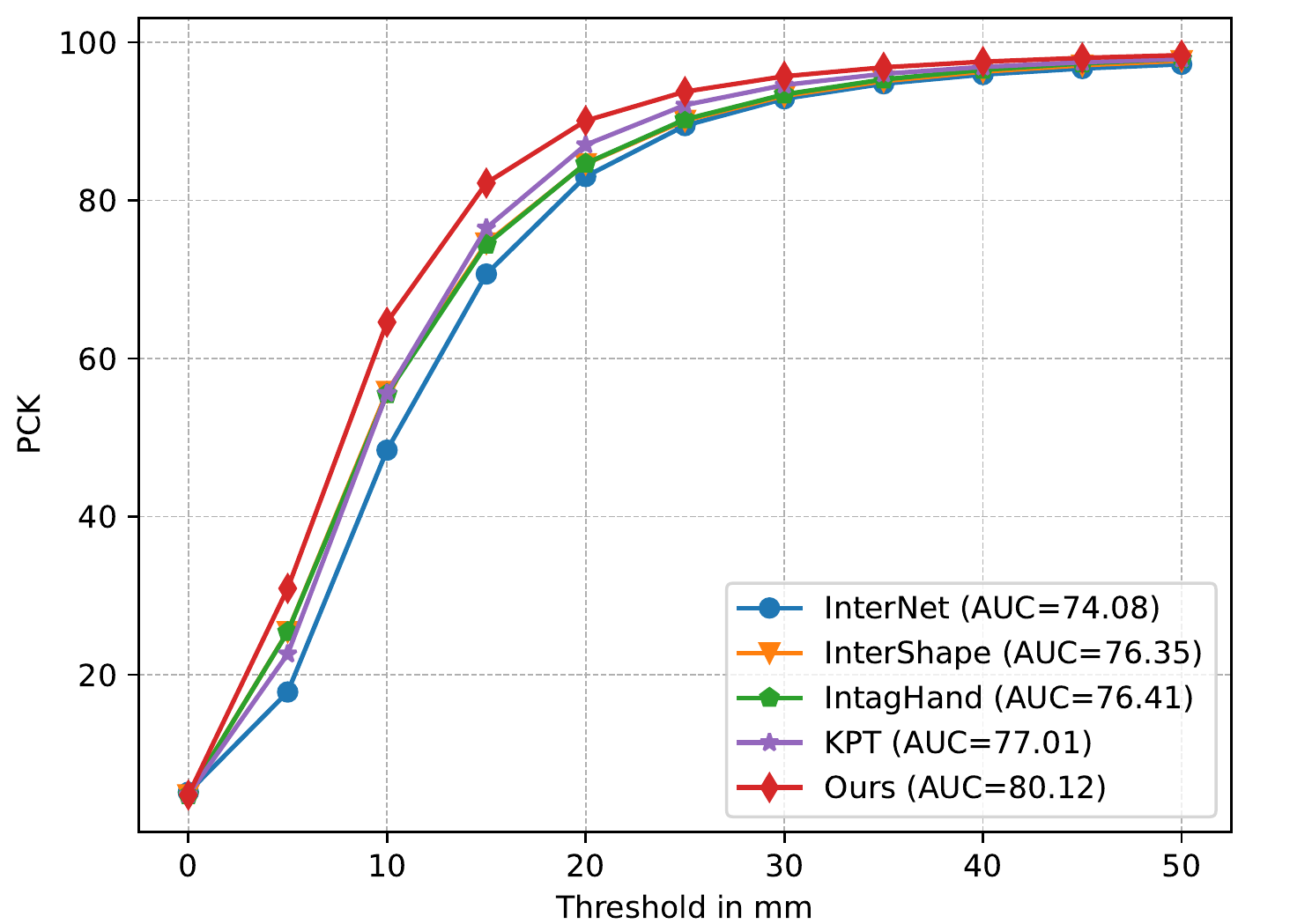}
\end{center}
  \caption{Comparison with SOTA methods on InterHand2.6M.}
    \label{fig:pck}
\end{figure}

\subsection{Comparisons with State-of-the-arts} We compare our method SOTA two-hand reconstruction methods, including InterNet \cite{moon2020interhand2}, InterShape \cite{zhang2021interacting}, IntagHand \cite{li2022interacting} and KeyPoint Transformer (KPT) \cite{li2022interacting}. For the fairness of comparison, we re-evaluate these methods in the same way, that is, using the ground-truth of wrist joint for hand alignment and scaling the estimated results according to the ratio of the predicted bone length to the ground-truth bone length. In particular, we evaluate these methods using their released source code and model weights. As shown in Table \ref{table:sota}, our method significantly outperforms all two-hand reconstruction methods. First, compared with SOTA two-hand estimation method IntagHand \cite{li2022interacting}, our method improves by 15.40\% (10.49 mm vs. 12.40 mm) and 15.14\% (10.26 mm vs. 12.09 mm) in MPJPE and MPVPE. Second, since our method explicitly models the two hand spatial relations in the 3D joint feature space, our method shows a significant advantage in MRRPE. Third, our method exhibits better pixel-aligned properties, achieving the lowest MIAA. In addition, as shown in Fig. \ref{fig:pck}, our method outperforms previous methods at almost all error thresholds and has the highest AUC.

\begin{table}[t]
\begin{center}
\setlength\tabcolsep{5pt} 
\begin{tabular}{l||cc||cc}
\hline
Method            & \begin{tabular}[c]{@{}c@{}}MPJPE\\ (MS)\end{tabular} & \begin{tabular}[c]{@{}c@{}}MPVPE\\ (MS)\end{tabular}  & \begin{tabular}[c]{@{}c@{}}MPJPE\\ (M)\end{tabular}  & \begin{tabular}[c]{@{}c@{}}MPVPE\\ (M)\end{tabular}  \\ \hline
InterNet  \cite{moon2020interhand2}        & 12.79  &  - & 12.94 &  - \\
KPT \cite{hampali2022keypoint}             & 9.17  & -  & 9.61 & - \\
InterShape \cite{zhang2021interacting}     & 8.88  & 9.15 & 10.51 & 10.95 \\
IntagHand  \cite{li2022interacting}        & 8.79  & 9.03 & 10.14 & 10.59 \\ 
\hline
Ours              & \textbf{7.51}   & \textbf{7.72} & \textbf{8.70} & \textbf{9.06} \\ \hline
\end{tabular}
\end{center}
\caption{We report MPJPE-MS (mm), MPVPE-MS (mm), MPJPE-M (mm), MPVPE-M (mm) on InterHand2.6M.}
\label{table:sota}
\end{table}

Meanwhile, we report MPJPE and MPVPE using the middle finger MCP joint for alignment while using scaling (the evaluation method is the same as IntagHand \cite{li2022interacting}), called MPJPE-MS and MPVPE-MS, respectively. We also report MPJPE and MPVPE using MCP alignment without scaling, referred to as MPJPE-M and MPVPE-M, respectively. As shown in Table \ref{table:sota}, evaluating with different settings can lead to significantly different performances. For example, using the middle finger MCP joint alignment can significantly improve the performance of the network. However, regardless of the evaluation strategy, our method outperforms SOTA methods by a large margin under all metrics. 

In addition, we adopt Interpenetration Volume (IV) to measure the degree of collision between two hands. IV is obtained by voxelising the hand model using a voxel size of 0.5 cm. Compared with previous SOTA method IntagHand \cite{li2022interacting}, without explicitly modeling the collision between the hands, our method has a lower IV (4.14 cm$^3$ Vs. 4.37 cm$^3$). This shows that our method better models the spatial relationship between hands.

\begin{figure}[ht]
  \centering
  \includegraphics[width=0.99\linewidth]{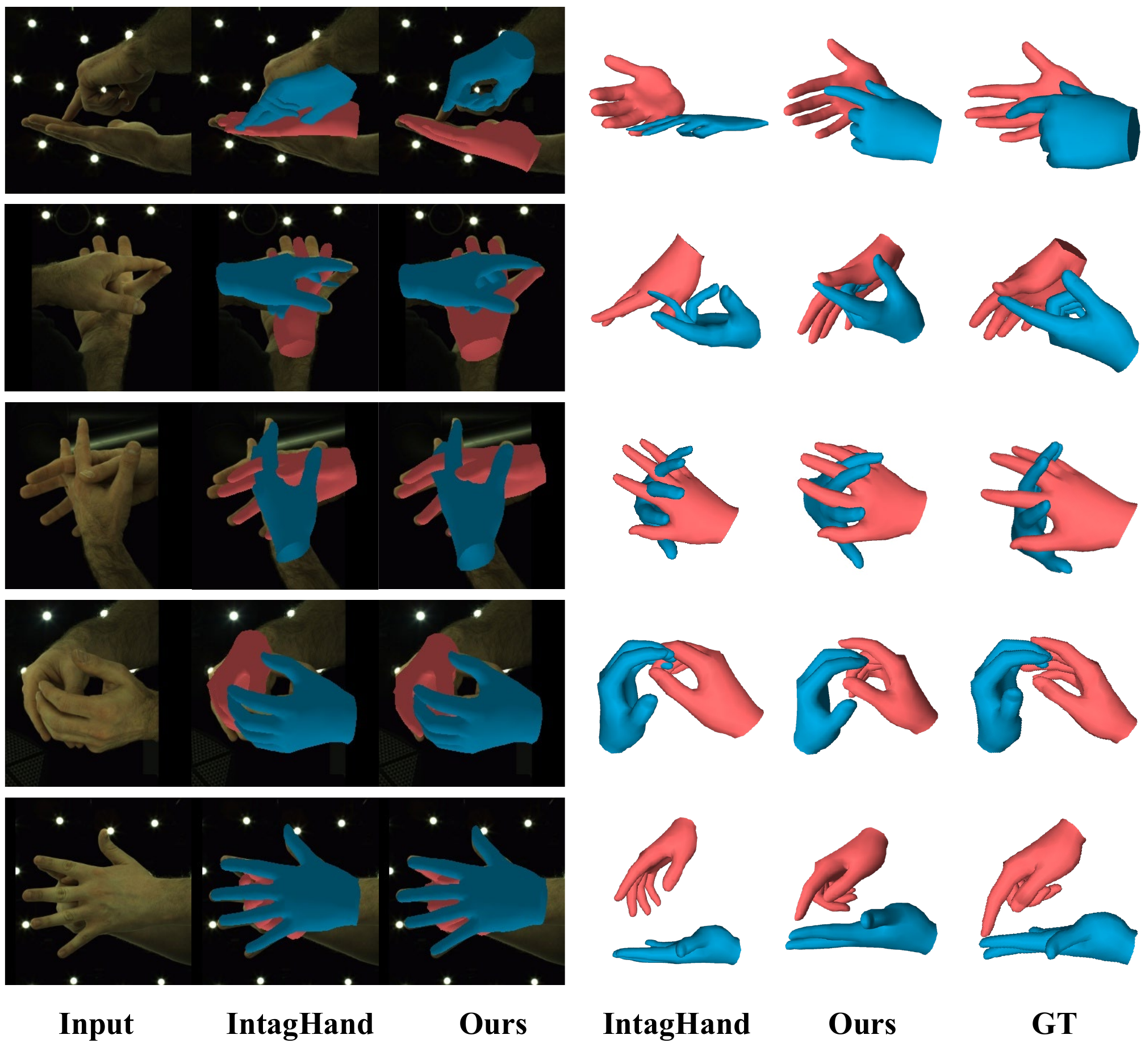}
\caption{Qualitative results of IntagHand \cite{li2022interacting} and our method on InterHand2.6M dataset. Left: frontal view to evaluate the alignment accuracy; Right: another view to observe the plausibility of the estimated two hand meshes. }
\label{fig:interhand}
\end{figure}

\subsection{Qualitative Results}
We present the qualitative results of our method on the Interhand2.6M in Fig. \ref{fig:interhand}.
Compared with IntagHand \cite{li2022interacting}, our method can avoid the collapse of the estimated mesh (row 1, row 2); without an explicitly collision constraint, our method can better avoid unreasonable intersections between hands (row 3, row 4). This shows that the joint-based information interaction can help our model capture the spatial relationship between the joints. Meanwhile, our method achieves better mesh-image alignment (row 2, row 3). In particular, when a hand is almost completely occluded (row 5), our method can also infer the pose of the occluded hand based on fine-grained visual cues and global information. We provide a comparison with IntagHand on live video in the supplementary material.

\begin{figure}[t]
\centering
   \includegraphics[width=0.99\linewidth]{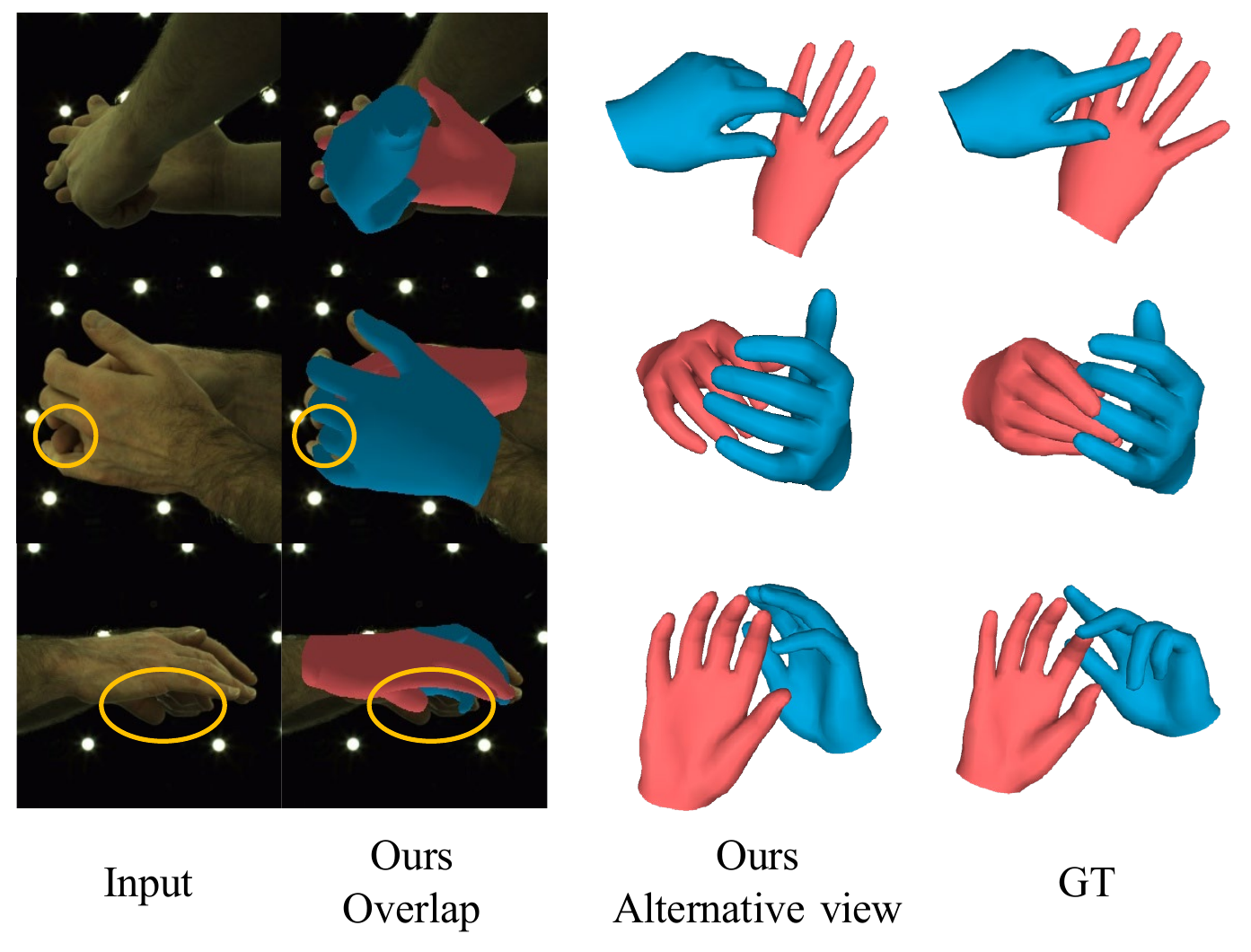}
   \caption{Failure examples on InterHand2.6M \cite{moon2020interhand2}. We highlight the region where the reconstruction is wrong with a yellow circle. `GT' means the ground truth.}
\label{fig:fail}
\end{figure}

We also show some failure examples of our method on the Interhand2.6M. As shown in Fig. \ref{fig:fail}, for cases with severe self-occlusion and fine-grained interactions (row1), our method does not achieve accurate mesh-image alignment and has a wrong understanding of the two-hand relationship. Second, for examples with severe self-occlusion, when the target joint has few observable pixels (row2) or the observable region is dark (row3), our method cannot reconstruct the corresponding region accurately. In conclusion, our method may fail when multiple conditions such as self-occlusion, tight interaction, blurring or shadowing occur simultaneously or in combination.

\begin{figure*}[t]
\begin{center}
   \includegraphics[width=0.99\linewidth]{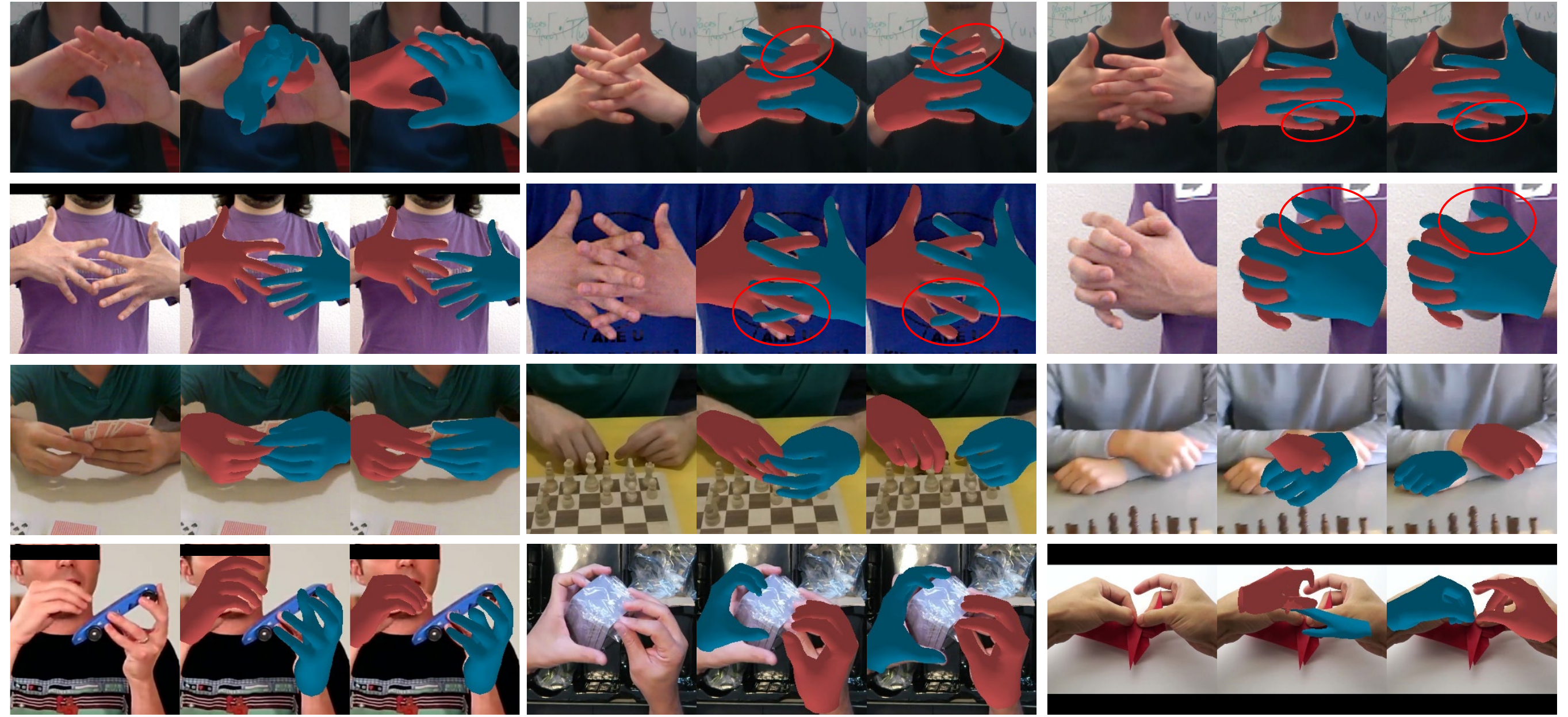}
\end{center}
  \caption{Qualitative results on in-the-wild images. Each row from top to bottom corresponds to RGB2Hands dataset\cite{wang2020rgb2hands}, the dataset proposed by Tzionas et al. \cite{tzionas2016capturing}, EgoHands dataset \cite{bambach2015lending} and 100DOH dataset \cite{shan2020understanding}.  In each part, the left is the input image, the middle is the result of IntagHand \cite{li2022interacting}, and the right is the result predicted by our method.}
\label{fig:wild}
\end{figure*}

Similar to \cite{li2022interacting, zhang2021interacting}, we also demonstrate the generalization ability of our method on in-the-wild images. It is worth mentioning that we only use InterHand2.6M dataset for training without any additional pre-training on synthetic hand datasets or fine-tuning on other hand datasets.
As shown in Fig. \ref{fig:wild}, our method has a strong generalization ability for different viewpoints such as third or egocentric
viewpoints. At the same time, as shown in the third row and the last row, our method is more robust to different backgrounds and lighting conditions. Since our method is able to enhance visual features with global information, our method is also relatively robust to object perturbations. In particular, compared with IntagHand \cite{li2022interacting}, which has strong generalization ability, our method also shows obvious advantages, especially in that our method can better maintain hand structure. We present more qualitative results in the supplementary material.

\section{Conclusion}
In this paper, we propose the decoupled iterative refinement framework to reconstruct interacting hands from a single RGB image.
In order to efficiently model the spatial dependencies between the two hands, we adopt the GCN and the transformer to perform intra- and inter-hand information interaction in the 3D joint feature space. 
To achieve better alignment of the estimated mesh and observed images, we project the joint features with global information into the visual feature space in a confusion-free manner, which provides strong disambiguation cues for visual features, alleviating self-occlusion and self-similarity problems. 
Ablation experiments demonstrate that the decoupled iterative refinement can effectively solve two major challenges in interacting hands reconstruction, namely,  modeling the complex hands spatial relationships and visual feature deconfusion.
Quantitative experiments on InterHand2.6M show that our method outperforms the previous SOTA by a large margin. Meanwhile, experiments on in-the-wild images demonstrate that our method has a strong generalization ability.

\textbf{Limitation and Future Work.}
Our method does not explicitly model collisions between hands, so even with the intra- and inter-hand relationships modeling, intersections between hands still occur, in some cases. Besides, our method does not fully utilize the estimated 3D mesh information. Mesh information may helpful for a fine-grained understanding of the relationship between hands. Finally, to achieve finer-grained mesh-image alignment, highfidelity parametric hand models may be beneficial.

\section*{Acknowledgement}
This work was supported in part by the National Natural Science Foundation of China under Grants (62071067, 62171057, 62201072), in part by the Ministry of Education and China Mobile Joint Fund (MCM20200202), Beijing University of Posts and Telecommunications-China Mobile Research Institute Joint Innovation Center, in part by the Project funded by China
Postdoctoral Science Foundation (2023TQ0039).

\maketitle
\appendix

\clearpage
\setcounter{section}{0}
\setcounter{table}{0}
\setcounter{figure}{0}
\begin{center}
\textbf{\Large Supplementary Materials}
\end{center}
\

In the supplemental material, we provide:

\begin{itemize}
\item[$\bullet $] more details of network structure and computational requirements in Sec. \ref{sec:network},
\item[$\bullet $] the details of mesh smooth loss in Sec. \ref{sec:loss},
\item[$\bullet $] more quantitative results in Sec. \ref{sec:quantitative},
\item[$\bullet $] more ablation experiments in Sec. \ref{sec:ablation},
\item[$\bullet $] qualitative results in real scenarios in Sec. \ref{sec:qualitative},
\end{itemize}

Note that all the notation and abbreviations here are consistent with the main manuscript.
\section{Details of Network Structure and Computational Requirements} \label{sec:network}

In this section, we introduce the structure details of the feature fusion layers, the Graph Convolutional Network (GCN) and the transformers. First, we use a residual convolutional module as the stacked hourglass network \cite{newell2016stacked} to fuse the feature maps from the encoder and previous decoding layer. Meanwhile, we use a residual convolution module to enhance the fused visual feature map with projected features. We set the number of channels of fused visual features and enhanced visual features to 256. We adopt a 4-layer semantic GCN \cite{zhao2019semantic} to perform information interaction between the single-hand joints, where the number of channels of joint features is 128. We adopt a 4-layer transformer \cite{zheng20213d} for information interaction between two-hand joints, in which we add spatial position encoding to the input joint features. 
Wtih a single GPU (NVIDIA A100) and a batch size of 64, for the network with two refinement stages, the training time is 39.8h, the memory usage is 22.1G, the FLOPs is 30.8G, and the model parameters are 55.1M.

\section{Mesh Smooth Loss} \label{sec:loss}
Following previous methods \cite{wang2018pixel2mesh, moon2020i2l, li2022interacting}, we use mesh smooth loss to maintain the estimated mesh geometry reasonable, including a normal consistency loss $L_{norm}$ and edge length consistency loss $L_{edge}$. $L_{norm}$ is defined as follows:
\begin{equation}
   L_{norm}=\sum_{f} \sum_{\left \{ i,j \right \} \subset f}\left \|   \left \langle \mathbf{e}_{ij}, \mathbf{n}_{f}^{gt} \right \rangle  \right \|_{1}, 
  \label{eq:normal}
\end{equation}
where $f$ and $\mathbf{n}_{f}$ indicate a face of the hand mesh and the unit normal vector of face $f$, respectively. $\mathbf{e}_{ij}$ indicates a edge of the face $f$. $\left \langle \cdot , \cdot \right \rangle$ is the inner product of two vectors.

$L_{edge}$ is defined as follows:
\begin{equation}
   L_{edge}=\sum_{f} \sum_{\left \{ i,j \right \} \subset f}\left \| \left \| \mathbf{e}_{ij} \right \|_{2}, \left \| \mathbf{e}_{ij}^{gt} \right \|_{2}  \right \|_{1}.
  \label{eq:edge}
\end{equation}
$L_{edge}$ constrains each edge of the predicted mesh to have the same edge length as the ground truth.

\begin{table}[b]
\begin{center}
\begin{tabular}{lcc}

\hline
               & MPJPE                  & MPVPE  \\ \hline
InterShape \cite{zhang2021interacting}     & 18.24                  & 17.93   \\ \hline
IntagHand \cite{li2022interacting}     & 19.21                 & 18.91   \\ \hline
Ours           & \textbf{16.48}          & \textbf{16.25}   \\ \hline
\end{tabular}
\end{center}
   \caption{Quantitative results on Tzionas et al. \cite{tzionas2016capturing}. We report the MPJPE (mm) and MPVPE (mm). }
   \label{table:tzionas}
\end{table}

\section{More Quantitative Results} \label{sec:quantitative}
In the dataset proposed by Tzionas et al. \cite{tzionas2016capturing}, we selected the sequence containing two hands for quantitative experiments. In particular, we only use this dataset for testing and all models are trained using the InterHand2.6M \cite{moon2020interhand2}. As shown in Table \ref{table:tzionas}, our method outperforms IntagHand \cite{li2022interacting} and InterShape \cite{zhang2021interacting} by a large margin on \cite{tzionas2016capturing}, which further demonstrates the superior generalization ability of our method.

\begin{table}[]
\begin{center}
\begin{tabular}{l|ccc}
\hline
Method          & MPJPE & MPJPE & MIAA \\ \hline
Baseline        & 12.44 & 12.11 & 7.41 \\ \hline
w/o Attn        & 12.53 & 12.23 & 7.53 \\
w/o Smooth L1   & 12.49 & 12.20 & 7.50 \\
w/o Motion Blur & 12.50 & 12.21 & 7.51 \\
w/o large LR    & 12.79 & 12.45 & 7.64 \\ \hline
w/o All         & 13.01 & 12.67 & 7.83 \\ \hline
\end{tabular}
   \caption{Ablation study of the basic model on InterHand2.6M \cite{moon2020interhand2}. We report the MPJPE (mm), MPVPE (mm) and MIAA (pixel).}
   \label{table:ablation}
\end{center}
\end{table}

\begin{figure*}
\centering
   \includegraphics[width=0.99\linewidth]{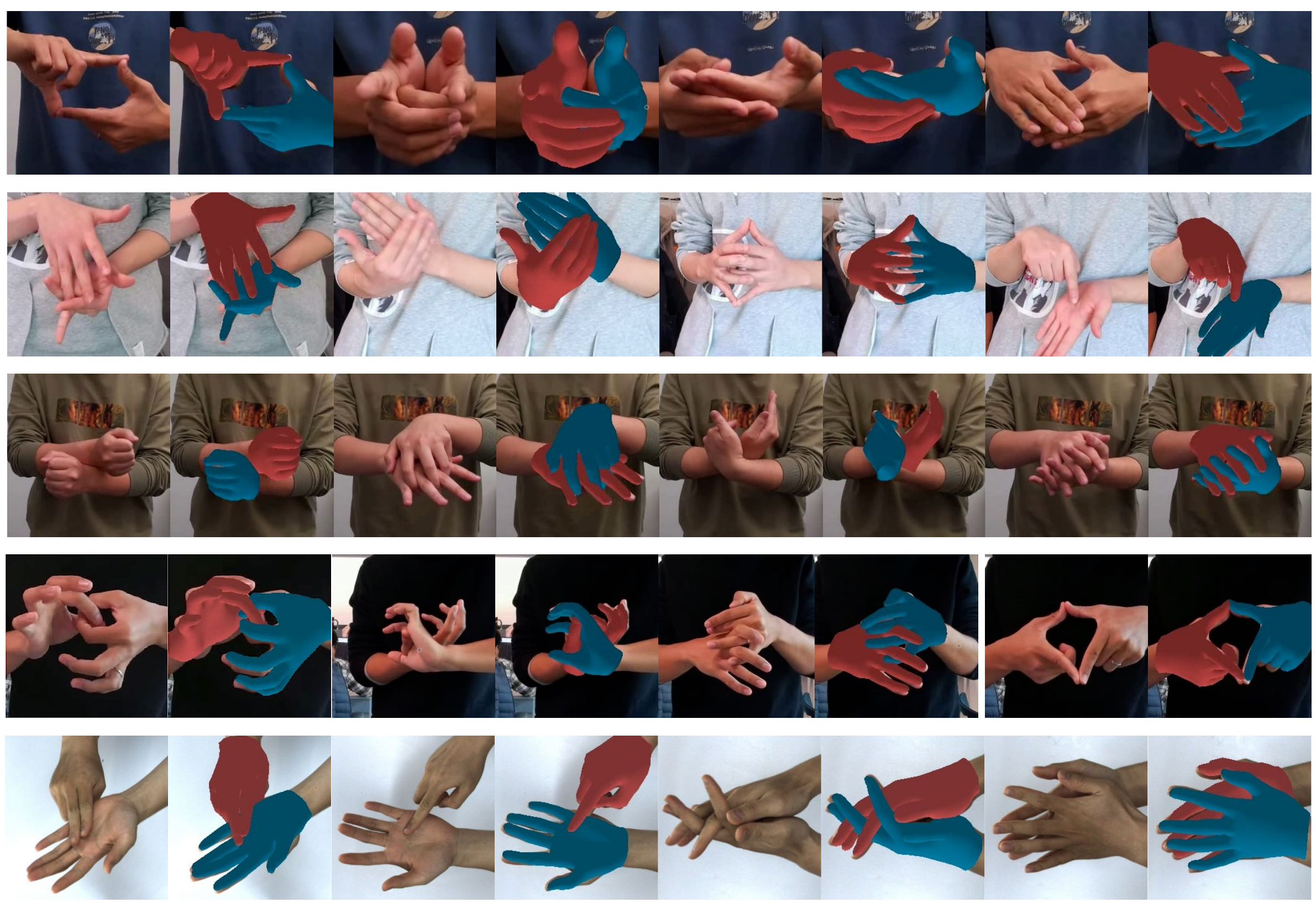}
   \caption{Two-hand reconstruction results in real scenarios on different subjects with different hand shapes and hand poses. }
\label{fig:wild_sup}
\end{figure*}

\section{Ablation Study} \label{sec:ablation}
Our basic model regresses the hand model parameters directly from the visual features. We tried multiple good practices to improve the performance of the basic model, including adopting a larger learning rate (from 1e-4 to 3e-4), adopting SmoothL1 loss \cite{girshick2015fast, ren2019srn} instead of L1 loss, adopting motion blur for data augmentation, adopting the attention mechanism to obtain the different feature of the left and right hand respectively, etc. As shown in Table \ref{table:ablation}, using the attention mechanism, data augmentation and Smooth L1 loss have an impact of about 0.1 mm on the basic model, and the use of a larger learning rate has an impact of close to 0.35 mm. If these components are removed, the MPJPE drops to 13.01 mm. It is worth mentioning that compared to the previous methods, our method can adopt a larger learning rate due to the simple and efficient network design.

\section{Qualitative Results} \label{sec:qualitative}
As shown in Fig. \ref{fig:wild_sup},  we experiment on five subjects in real scenarios.  The five subjects have different hand shapes and hand poses. First, our method is able to generate relatively accurate mesh-image alignments for unseen subjects. Second, our method can also perform reasonable reconstructions for some unseen complex interacting poses. Overall, our method achieves efficient pixel-level alignment and 3D spatial relationship modeling thanks to the decoupled design of 2D visual feature space and 3D pose feature space. At the same time, sparse and compact node-level information interaction avoids overfitting and achieves strong generalization ability. In particular, we provide a video in the Supplementary Materials to demonstrate the strengths of our method for spatial relationship modeling and image-mesh alignment compared to SOTA method IntagHand \cite{li2022interacting}.

\newpage

{\small
\bibliographystyle{ieee_fullname}
\bibliography{egbib}
}

\end{document}